\documentclass[10pt,twocolumn,letterpaper]{article}

\usepackage{iccv}
\usepackage{array}
\usepackage{arydshln}
\usepackage{times}
\usepackage{epsfig}
\usepackage{graphicx}
\usepackage{amsmath}
\usepackage{amssymb}
\usepackage{dsfont}
\usepackage{multirow}
\usepackage{cite}

\usepackage{enumitem}
\usepackage{caption}
\usepackage{algorithm,algpseudocode}
\usepackage{color}
\usepackage{booktabs}
\usepackage{bbding}

\usepackage[breaklinks=true,bookmarks=false, colorlinks,linkcolor=red,anchorcolor=red,citecolor=blue]{hyperref}

\iccvfinalcopy 

\def\iccvPaperID{5056} 

\ificcvfinal\fi

\makeatletter
\newcommand{\thickhline}{%
    \noalign {\ifnum 0=`}\fi \hrule height 1pt
    \futurelet \reserved@a \@xhline
}
\newcolumntype{"}{@{\hskip\tabcolsep\vrule width 1pt\hskip\tabcolsep}}
\makeatother

\DeclareMathOperator*{\argmax}{arg\,max}  

\begin{document}

\title{Learning Fine-Grained Features for Pixel-wise Video Correspondences}

\author{Rui Li ~~~~~~~ Shenglong Zhou ~~~~~~~ Dong Liu  \\
         University of Science and Technology of China, Hefei, China\\
{\tt\small \{liruid,slzhou96\}@mail.ustc.edu.cn, \tt\small dongeliu@ustc.edu.cn}\\
\small\textbf{\url{https://github.com/qianduoduolr/FGVC}}
}

\maketitle
\ificcvfinal\thispagestyle{empty}\fi

\begin{abstract}
   \footnotetext[1]{ This work was supported by the Natural Science Foundation of China under Grants 62022075 and 62036005, and by the Fundamental Research Funds for the Central Universities under Grant WK3490000006. This work was also supported by the advanced computing resources provided by the Supercomputing Center of USTC. \emph{(Corresponding author: Dong Liu.)}}
   Video analysis tasks rely heavily on identifying the pixels from different frames that correspond to the same visual target. To tackle this problem, recent studies have advocated feature learning methods that aim to learn distinctive representations to match the pixels, especially in a self-supervised fashion. Unfortunately, these methods have difficulties for tiny or even single-pixel visual targets. Pixel-wise video correspondences were traditionally related to optical flows, which however lead to deterministic correspondences and lack robustness on real-world videos. We address the problem of learning features for establishing pixel-wise correspondences. Motivated by optical flows as well as the self-supervised feature learning, we propose to use not only labeled synthetic videos but also unlabeled real-world videos for learning fine-grained representations in a holistic framework. We adopt an adversarial learning scheme to enhance the generalization ability of the learned features. Moreover, we design a coarse-to-fine framework to pursue high computational efficiency. Our experimental results on a series of correspondence-based tasks demonstrate that the proposed method outperforms state-of-the-art rivals in both accuracy and efficiency.
\end{abstract}

\section{Introduction}

One of the most fundamental problems in computer vision is learning visual correspondences across space and time, which has many applications such as 3D reconstruction, physical understanding, and dynamic object modeling. Due to the factors such as viewpoint change, distractors, and deformations, this task is extremely challenging and can be roughly divided into three categories according to the granularity: the first one is object-wise correspondences that exist between coarsely localized bounding boxes~\cite{wang2019unsupervised, valmadre2017end} along the video; the second one is group-wise correspondences, indicating the mapping at group-level, and usually applied to downstream tasks like video object segmentation~\cite{caelles2017one, maninis2018video}; the third one is pixel-wise correspondences, which describe the pixel-level relation between video frames with the finest granularity. 
\begin{figure}[!t]
   \centering
   {\includegraphics[width=0.48\textwidth]{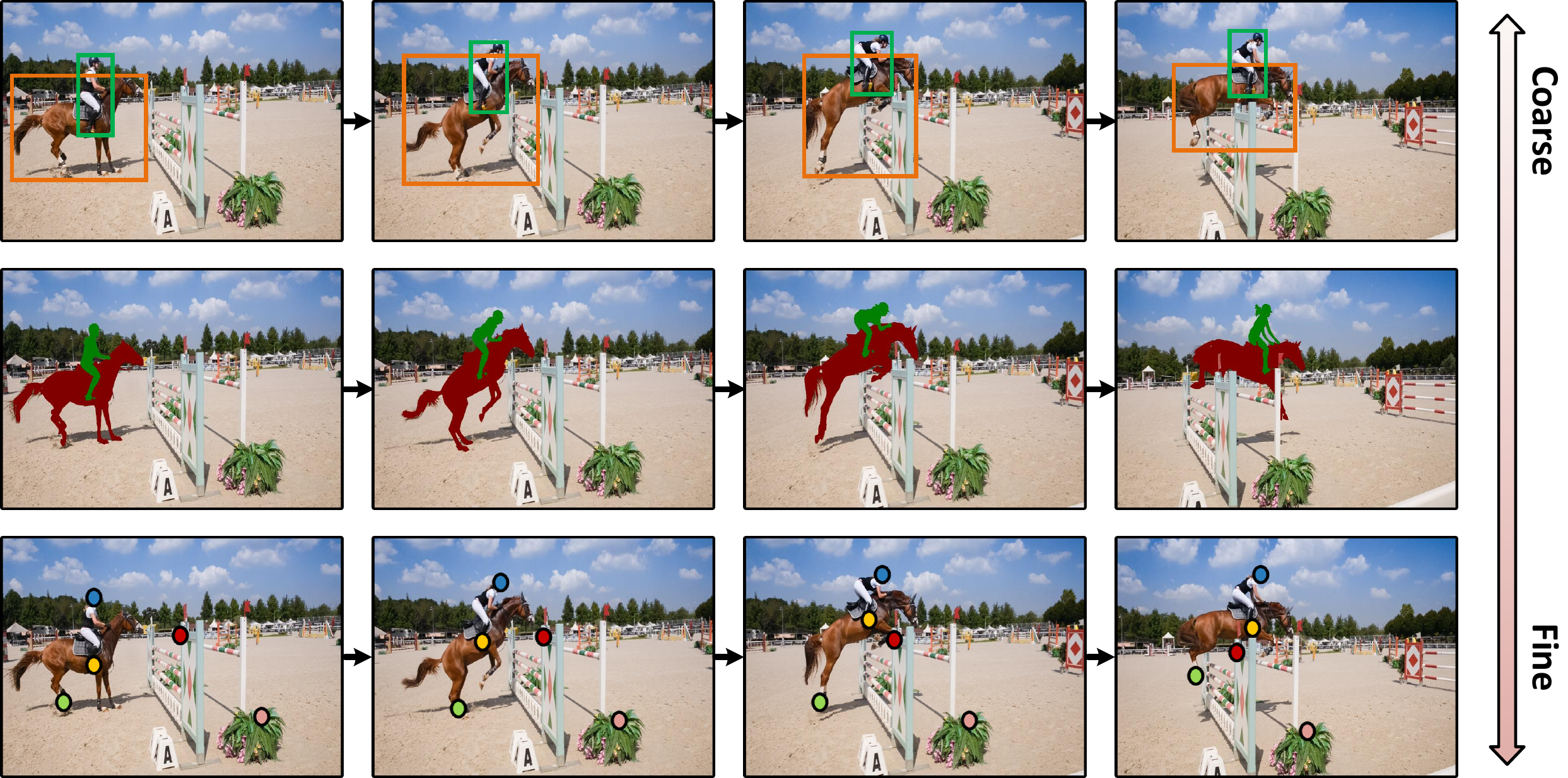}}
   \vspace{-6mm}
   \caption{\small We illustrate video correspondences with different granularities, including object-wise, group-wise,  and pixel-wise. In this paper, we concentrate on learning fine-grained features to address the pixel-wise video correspondences. }
   \label{fig:teaser}
   \vspace{-5mm}
\end{figure}

Learning dense representations from videos is one approach to finding the correspondences. Researchers have been exploring different self-supervised methods for learning generalizable representations using unlabeled videos collected in the real world~\cite{jabri2020space,wang2019learning,xu2021rethinking,lai2020mast,li2023spatial,li2019joint}. For example, Wang et al.~\cite{wang2019learning} propose using an object-level cycle-consistency across time (i.e., forward-backward tracking) as a supervision signal. Jabri et al.~\cite{jabri2020space} further enhance it by combining cycles of time with the similarities between path-level representations. Inspired by contrastive learning, Xu et al.~\cite{xu2021rethinking} try to learn spatial and temporal representation through a frame-wise contrastive loss, while Li et al.~\cite{li2023spatial} propose a spatial-then-temporal pretext task to learn better spatiotemporal features.  Despite obtaining promising outcomes, current research has predominantly emphasized performing object-level or patch-level similarity learning, making it difficult to accurately recognize pixel-wise differences with the learned features. As a result, there is an increasing necessity for learning fine-grained representations in order to tackle this problem.

At the same time, there is another line of work approaching video correspondences by deterministically predicting the displacement of each pixel, which is known as optical flow estimation. Early studies used optimization methods to estimate the motion between two frames~\cite{brox2010large}. In recent years, approaches use synthetic data with supervised learning for flow estimation~\cite{dosovitskiy2015flownet,mayer2016large}, using a coarse-to-fine pyramid framework to improve the accuracy~\cite{sun2018pwc}. RAFT~\cite{teed2020raft} further devises an iterative optimization algorithm to come up with the result of high-resolution flow fields through iterative updates, which show a superior ability to find fine-grained correspondences. However, in real scenarios, there are often appearance variants, illumination changes, and deformations between video frames, which leads to the lack of robustness on real-world videos for the optical flow model supervised by labeled synthetic videos.

In this paper, we explore how to learn fine-grained representations to meet the needs of pixel-wise video correspondences. To this end, we first investigate how to leverage synthetic data for fine-grained feature learning.  Specifically, given a query pixel, the supervision in synthetic videos only supplies the one-to-one mapping, i.e., a motion vector, representing the deterministic correspondence of the pixel to another pixel in the next frame. Nevertheless, the pixel-wise features evolve slowly over space and time, which indicates a soft distribution of the correspondences. We find directly utilizing the synthetic supervision as hard labels results in inferior representations, and the learned features are unable to recognize the pixel-wise differences across different spatial locations and periods of time. To address the issue, we propose to use an external pre-trained 2D encoder to derive soft supervision for optimization based on the flow. 

Furthermore, we incorporate self-supervised feature learning on unlabeled real data in the overall training to alleviate the generalization issues in real scenes, which consist of two carefully designed components. Firstly, inspire by the temporal consistency assumption~\cite{black1993framework}, we learn temporal persistent features via self-supervised reconstructive learning, where each query pixel can be reconstructed by leveraging the information in adjacent frames. Besides, given the synthetic and real data, we perform adversarial training by introducing Gradient Reverse Layer~\cite{ganin2015unsupervised} with a discriminator for the learned correspondences. We observe such designs can further enhance learned fine-grained features. 

Though already getting impressive results, we find it takes more time to get the results of the dense matching between fine-grained features. Thus, we make another step to devise a coarse-to-fine framework to address the problem. We put the complex feature matching on the coarse-grained feature map and then get the fine-grained results through a learnable up-sampling layer. As a result, we achieve a good balance of performance and efficiency.
In summary, the main contribution of this work lies in:

\begin{itemize}[leftmargin=*]
   \item We address the problem of establishing pixel-wise video correspondences via a feature learning approach.
   \item We propose an effective method of learning fine-grained features from both synthetic and unlabeled videos, followed by a carefully designed framework to address the issue of efficiency.
   \item We validate our method in a series of correspondence-based tasks. Experiment results indicate consistent improvement over state-of-the-art methods.
\end{itemize}

\section{Related work}
\textbf{Representation learning for video correspondences.} 
Recent researches center around learning dense representations without labels in a self-supervised way for video correspondences, which occurs in two distinct directions: reconstruct-based ~\cite{lai2019self, lai2020mast, li2023spatial, li2019joint, vondrick2018tracking, wang2020contrastive} and cycle-consistency-based techniques~\cite{jabri2020space, wang2019learning, zhao2021modelling}. In the works of the first type, the query pixel is reconstructed from the adjacent frames based on the temporal consistency assumption, while the works of the second type execute forward-backward tracking to reduce cycle inconsistency. Furthermore, VFS~\cite{xu2021rethinking} proposes to learn representations through frame-wise contrastive loss. SFC~\cite{hu2022semantic} proposes a two-stream architecture to learn semantic and fine-grained features through two different models. Despite the progress made in learning representations for video correspondences, accurately recognizing pixel-wise distinctions over space and time remains challenging. 

\begin{figure*}[t]
   \centering
   {\includegraphics[width=0.85\textwidth]{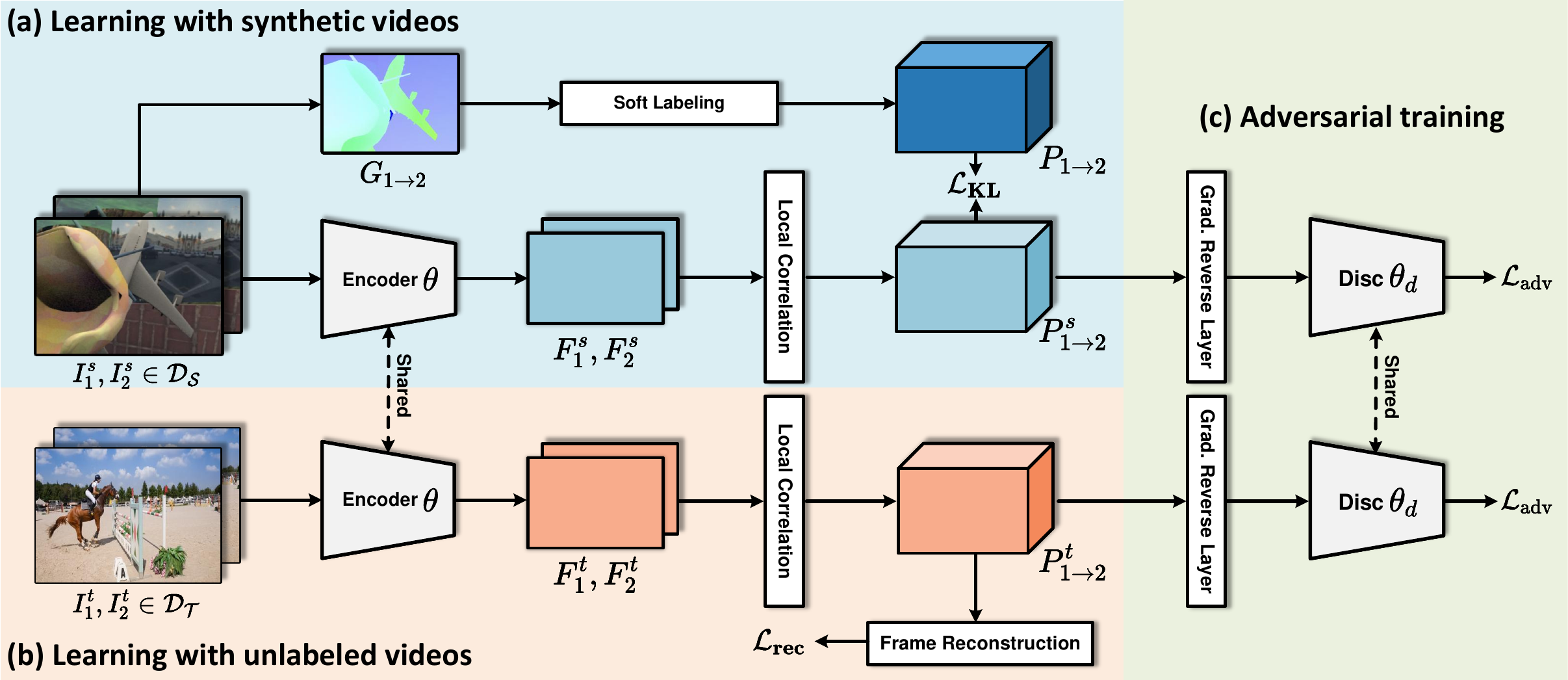}}
   \caption{\small \textbf{Overview of the framework for fine-grained feature learning.} We facilitate fine-grained feature learning by integrating self-supervised learning on unlabeled videos into supervised learning with labeled synthetic videos. For learning with synthetic data $\mathcal{D}_{\mathcal{S}}$, we devise a soft labeling module to convert the hard labels indicated by the motion vectors to soft labels. To learn more generalizable features, we leverage free supervision from unlabeled videos  $\mathcal{D}_{\mathcal{T}}$ with the objective of frame reconstruction. Furthermore, an adversarial loss~(along with Gradient Reverse Layer and a discriminator) is proposed to encourage domain invariant representations. The whole framework is jointly optimized in an end-to-end manner with the proposed objective functions. }
   \label{fig:framework}
   \vspace{-6mm}
 \end{figure*}

\textbf{Optical flow estimation for video correspondences. } 
Recently, the classic optical flow estimation problem of predicting per-pixel motion between two frames has been explored using synthetic graphics data for supervised training~\cite{dosovitskiy2015flownet,mayer2016large}. FlowNet~\cite{dosovitskiy2015flownet} was one of the first deep learning methods to tackle end-to-end optical flow learning. This research inspired a multitude of other methods, such as FlowNet2.0~\cite{ilg2017flownet}, DCFlow~\cite{xu2017accurate}, SpyNet~\cite{ranjan2017optical}, PWC-Net~\cite{sun2018pwc}, and LiteFlowNet3~\cite{hui2020liteflownet3}. Most of these methods employ cost volumes for finding pixel matching. RAFT~\cite{teed2020raft} stands out from the rest due to its multi-scale correlation volumes and iterative flow refinements, whilst achieving superior performance, and is also a precursor to many successive works. However, learning a deterministic model with synthetic computer graphics data limits generalization ability and robustness on real videos.

\textbf{Unsupervised domain adaptation with self-supervised learning.} Recently, there has been a surge in approaches to reduce the distribution discrepancy between real and synthetic data by leveraging unsupervised domain adaptation~\cite{ganin2015unsupervised,kang2019contrastive,pan2019transferrable,xiao2021dynamic,zhang2018fully,tzeng2017adversarial}. 
An effective way to realize it is through adversarial training. Recent studies execute adversarial training by learning a domain classifier (the discriminator) to distinguish the learned features from different distributions, utilizing adversarial loss to increase domain confusion. Meanwhile, self-supervised learning obtains impressive performance and shows good generalizing ability by designing different pretext tasks~\cite{he2020momentum,  wang2019learning,xu2021rethinking,xie2021detco} with unlabeled data, which motivates us to combine supervised learning and self-supervised learning to encourage consistent representations for both domains as well as improved results for downstream tasks.

\section{Approach}
We address the problem of estimating the pixel-wise correspondences between a pair of video frames, which can be realized by learning fine-grained features for matching. Our goal is to learn a fine-grained feature space $\phi$ with the encoder~$\theta$ by designing different learning objective functions for the probabilistic mapping.

\textbf{Probabilistic mapping.} Given a pair of video frames~$I_1, I_2  \in \mathbb{R}^{h \times w \times 3}$, for all pixels in $I_1$,   we aim to predict the probabilistic mapping $P_{1\rightarrow2} \in \mathbb{R}^{H \times W \times (2r+1)^2}$, and $P_{1\rightarrow2}(\cdot|i) \in \mathbb{R}^{(2r+1)^2}$ gives the probability that $i$ is mapped to $j$ in frame $I_2$ within a limited range $r$, considering the nature of temporal coherence in the video. The $i,j \in \mathbb{R}^2 $ indicate the 2D pixel location. $P_{1\rightarrow2}(\cdot|i) \in \mathbb{R}^{(2r+1)^2}$ thus encodes the entire discrete conditional probability distribution of where $i$ is mapped in frame $I_2$. The probabilistic mapping can be achieved by calculating the feature similarities using the learned fine-grained features. More specifically,  we first extract the dense features~$F_1, F_2 \in \mathbb{R}^{H \times W \times C}$. Then the discrete probabilistic map can be obtained by computing the local correlation w.r.t. each key $j$ in $I_2$ within a local window for each query $i$,
\begin{equation}\label{eq:local_correlation}
  \footnotesize
  \begin{aligned}
     P_{1\rightarrow2}(j|i)=\frac{\exp \left(F_1(i)  \cdot F_2(j) / \tau\right)}{\sum_{n} \exp \left(F_1(i) \cdot F_2(n) / \tau\right)}, i \in \{1,.,HW\}, j,n \in \mathcal{K}(i)
  \end{aligned}~~,
\end{equation}
where $\mathcal{K}(i)$ is the index set in the local window with a limited range of $r$ centered at $i$, and $\tau$ is the temperature. The result of the probabilistic mapping is further post-processed and directly applied to various downstream tasks.

\subsection{Fine-Grained Feature Learning}\label{fine_grained_learning}
We design the learning objective functions for probabilistic mapping with both synthetic and real-world videos. The overview of the framework is shown in Figure~\ref{fig:framework}. 

\begin{figure}[t]
   \centering
   {\includegraphics[width=0.47\textwidth]{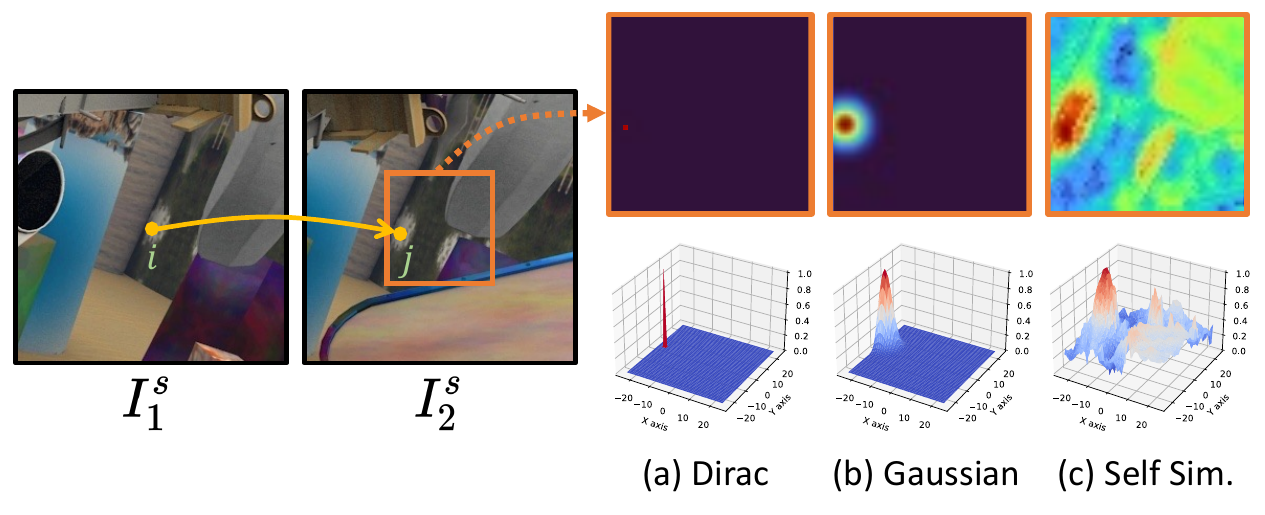}}
   \vspace{-3mm}
   \caption{\small \textbf{Illustration of the probabilistic map.} For $i$ in $I^s_1$, we visualize the probability value of the mapping from $i$ to the locations in  $I^s_2$ within a limited range. }
   \label{fig:self_labeling}
   \vspace{-6mm}
   \end{figure}

\textbf{Learning with synthetic videos.} Labeled synthetic videos are often used as supervision for learning the optical flows in recent studies~\cite{dosovitskiy2015flownet,teed2020raft}. Given a pair of rendered video frames~$I^s_1, I^s_2$, the synthetic data~$\mathcal{D}_{\mathcal{S}}$ provides pixel-wise motion vector, i.e., optical flow~$G_{1\rightarrow2}$. A valid question then emerges as how to learn features using such deterministic correspondences? Indeed, we believe the deterministic correspondences are hard to obtain for real-world videos, thus we argue the necessity to convert them into soft (probabilistic) ones. We devise the following variants:

\vspace{1mm}
\noindent (i)~\emph{Dirac distribution}: As shown in Figure~\ref{fig:self_labeling}~(a), we can directly convert the motion vector to a dirac distribution~$\delta(\cdot|i) $ in order to describe the ground truth mapping. Then, the learning objective function defined as Kullback-Leibler divergence between $P^s_{1\rightarrow2}(\cdot|i)$ and $\delta(\cdot|i)$. The $P_{1->2}^s$ stands for the correlation calculated by Eq.~(\ref{eq:local_correlation}) with synthetic data:

\begin{equation}\label{eq:klv1}
   \mathcal{L}_{\text{KL-v1}}=\sum_i D_{K L}(\delta(\cdot|i) ~\|~ P^s_{1\rightarrow2}(\cdot|i))
   \end{equation}

\noindent (ii)~\emph{Gaussian distribution}: However, the feature of the query pixel $i$ varies smoothly over space and time, which indicates a soft distribution of the potential location in the next frame. As shown in Figure~\ref{fig:self_labeling}~(a), the dirac distribution does not provide the feature learning with any knowledge of relative probability in the background, making it hard to learn the features with the ability to tell the fine-grained differences with synthetic videos. Thus, we devise to generate the probabilistic map with gaussian distribution, which introduces a soft distribution centered at the ground truth coordinate:
\begin{equation}
   \small
   \mathcal{N}\left(j|i, \mu, \sigma^2\right)=\frac{1}{\sqrt{2 \pi}\sigma_u} e^{-\frac{\left(u_j-\mu^u_i\right)^2}{2\sigma_u^2}} \cdot \frac{1}{\sqrt{2 \pi}\sigma_v} e^{-\frac{\left(v_j-\mu^v_i\right)^2}{2\sigma_v^2}},
   \end{equation}
where $\mu = (\mu^u_i,\mu^v_i)$ represents the 2D coordinate of ground truth in the $I^s_2$ for pixel~$i$, and  $(u_j,v_j)$ indicates the  2D coordinate of $j$. Then the loss can be devised as:

\begin{equation}\label{eq:klv2}
   \mathcal{L}_{\text{KL-v2}}=\sum_i D_{K L}(\mathcal{N}(\cdot|i, \mu, \sigma^2) ~\|~ P^s_{1\rightarrow2}(\cdot|i)).
   \end{equation}

\noindent (iii)~\emph{Soft-labeling}: The gaussian prior only considers the euclidean distance of coordinates, which shows limited capability of modeling the complicated distributions. We believe the synthetic supervision already provides valuable temporal cues~($i$ in $I^s_1$ move to $j$ in $I^s_2$), and the soft distribution over space can be approached by computing the self-similarities for $j$ using the pre-trained 2D visual encoder $\theta_{self}$ which can produce spatially-discriminative features. As shown in Figure~\ref{fig:self_labeling}~(c), for the $i$ in $I^s_1$ that moves to $j$, we get the feature of the query $i$~(denoted as $\overline{F}_q$) at the location $j$ in $\overline{F}_2$~($\overline{F}_2$ is the feature map of $I^s_2$ computed by~$\theta_{self}$). Then we can compute the feature similarities $S_{1\rightarrow2}(\cdot|i)$  between $\overline{F}_q$ and $\{\overline{F}_2(k)| k \in \mathcal{K}(i) \}$, where $\mathcal{K}(i)$ is the index set in the local window centered at $i$. The $S_{1\rightarrow2}(\cdot|i)$ are further normalized (by softmax) to obtain the discrete probability distribution $P_{1\rightarrow2}(\cdot|i)$:
\vspace{-3mm}

\begin{equation}\label{eq:klv3}
   \mathcal{L}_{\text{KL-v3}}=\sum_i D_{K L}(P_{1\rightarrow2}(\cdot|i) ~\|~ P^s_{1\rightarrow2}(\cdot|i)).
   \end{equation}

In Table~\ref{tab:abalation_kl}, we find soft labeling works well when only pre-training~$\theta_{self}$ on synthetic data~$\mathcal{D}_{\mathcal{S}}$ with $\mathcal{L}_{\mathrm{rec}}$~(Eq.~(\ref{eq:self_loss})), and leveraging  more strong 2D encoder would contribute to better performance. The comparisons between different loss functions will be discussed in the experiments, and the $\mathcal{L}_{\text{KL-v3}}$ is used as the default loss.

\textbf{Learning with unlabeled videos.} Meanwhile, we observe in real scenarios, there are apparent differences with synthetic videos in appearance variants, illumination changes and deformations, leading to notable changes in the distribution, where the learned features on synthetic videos show unsatisfied generalization ability. Inspired by recent studies~\cite{wang2019learning,jabri2020space}, we try to improve the learned fine-grained features by introducing self-supervised feature learning into the framework. As observed in the bottom of Figure~\ref{fig:teaser}, the pixel repetition encourages us to learn the fine-grained features by reconstructive learning, where each pixel in the ~$I^t_1$ can be reconstructed by leveraging the information of ~$I^t_2$ with a limited range. To achieve this, the video frames $I^t_1,I^t_2$  are firstly projected into pixel embeddings $F^t_1, F^t_2$ by the encoder $\theta$.   For each query $i$ in $I^t_1$,  we calculate the probabilistic map $P^t_{1\rightarrow2}$ with Eq.~(\ref{eq:local_correlation}). Then the query $i$ in  $I^t_1$ can be reconstructed by a weighted sum of pixels in $\mathcal{K}(i)$:
\begin{equation}\label{eq:reconstruction}
  \small
  \begin{aligned}
   {I}^{rec}_1(i)=\sum_{j \in \mathcal{K}(i)} P^t_{1\rightarrow2}(j|i) I^t_2(j).
  \end{aligned}~~
\end{equation}
 Then the reconstruction loss for self-supervised training is defined as $L_1$ distance between ${I}^{rec}_1$ and $I^t_1$. Training with such self-supervision leads to temporal persistent features that generalize well in real scenarios.

 However, the pixel repetition does not hold for pixels that become occluded, Thus, we exclude occluded pixels from the reconstruction loss to avoid learning incorrect features. We follow the forward-backward consistency assumption to detect the occluded pixels. which is defined in Eq.~(\ref{eq:occ}) as the occlusion flag $O_{1 \rightarrow 2}$ to be 1 whenever the constraint is violated, and 0 otherwise:
\begin{equation}\label{eq:occ}
   \small
   O_{1 \rightarrow 2}(i)=\mathds{1}\left(\argmax_{i} P^t_{2 \rightarrow 1}\left(i~|~\argmax_jP^t_{1 \rightarrow 2}(j~|~i)\right)=i\right)
   \end{equation}
The loss for reconstructive learning is defined as follows:
\begin{equation}\label{eq:self_loss}
   \small
   \begin{aligned}
      \mathcal{L}_{\text{rec}}= \sum_{i} O_{1 \rightarrow 2}(i)\cdot\left\|{I}^{rec}_1(i) - I^t_1(i)\right\|_{1}
   \end{aligned}~~.
\end{equation}

\textbf{Adversarial training.}  We further improve the fine-grained features by leveraging the technique in recent works of unsupervised domain adaptation, which aims to bridge the gap caused by the domain shift between the synthetic videos~$\mathcal{D}_{\mathcal{S}}$ and real videos~$\mathcal{D}_{\mathcal{T}}$ via adversarial training. An effective way to approach this problem consists in introducing the network a Gradient Reversal Layer~(GRL). We additionally train a discriminator~$\theta_D$ to identify whether the probabilistic map comes from synthetic  or real videos:
\begin{equation}\label{eq:da_loss}
   \small
   \mathcal{L}_{\text{adv}}= \mathbb{E}_{P^t \in P_{\mathcal{T}}}\left[\log \left(D(P^t)\right)\right]
    +~~\mathbb{E}_{P^s \in P_{\mathcal{S}}}\left[\log \left(1 - D(P^s)\right)\right],
   \end{equation}
and then we reverse the gradient direction during the backward pass in back-propagation when updating the parameters of the encoder~$\theta$ with $\mathcal{L}_{\text{adv}}$:

\begin{equation}
   \theta \longleftarrow \theta+\lambda \frac{\partial \mathcal{L}_{\text{adv}}}{\partial \theta} 
   \end{equation}

The GRL allows to train the discriminator and the encoder at the same time, and the adversarial training helps to learn domain invariant patterns.

\textbf{Overall training objective.} The overall training objective for learning fine-grained features is formulated as a multi-task loss, which is written as $ \mathcal{L} = \mathcal{L}_{\text{KL}} + \mathcal{L}_{\text{rec}} + \mathcal{L}_{\text{adv}}$, where we empirically treat each loss equally.

\begin{figure}[t]
   \centering
   {\includegraphics[width=0.48\textwidth]{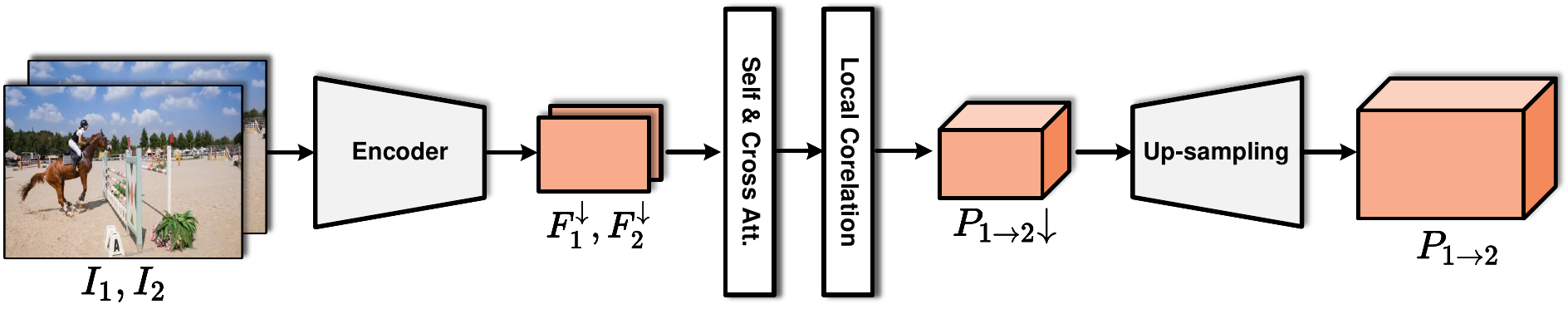}}
   \caption{\small \textbf{Illustration of the coarse-to-fine framework for efficient probabilistic mapping.} We first obtain coarse-grained matching and then upsample it to get the fine-grained result.}
   \label{fig:corrup}
   \vspace{-6mm}
 \end{figure}

\subsection{Efficient Fine-grained Probabilistic Mapping}
To get the fine-grained probabilistic map, i.e., $P_{1\rightarrow2} \in \mathbb{R}^{H \times W \times (2r+1)^2}$, we need to compute the similarities w.r.t. all key pixels in the local window centered at every query pixel, which is computationally costly. In this section, we devise a coarse-to-fine framework. The overview of the framework can be found in Figure~\ref{fig:corrup}, where we first compute the correlation at coarse-grained features and then upsample it to get fine-grained probabilistic maps.

\textbf{Self-attention and Cross-attention Layers.}   More specifically, we first extract the coarse feature maps $F^{\downarrow}_1,  F^{\downarrow}_2 \in \mathbb{R}^{H/4 \times W/4 \times C}$, and we further enhance the coarse features by introducing the self-attention and cross-attention layers with positional encoding. Then, we obtain the coarse-grained probabilistic map~$P_{1\rightarrow2}{\downarrow} \in \mathbb{R}^{H/4 \times W/4 \times (2r\downarrow + 1)^2}$ by Eq~(\ref{eq:local_correlation}).

\textbf{Up-sampling.} We devise an up-sampling layer to obtain fine-grained probabilistic map $P_{1\rightarrow2} \in \mathbb{R}^{H \times W \times (2r+1)^2}$, which can be simply done by leveraging pixel-shuffle or convolution layer with bilinear interpolation. 

More details of the architecture are included in supplementary material. We train the coarse-to-fine framework in a distillation manner on $\mathcal{D}_{\mathcal{T}}$, with the same objective function as $\mathcal{L}_{\text{KL}}$. The supervisions for $P_{1\rightarrow2}$ are obtained by computing the probabilistic map using pre-trained encoder in Sec.~\ref{fine_grained_learning}. Without losing much performance, we find such a design helps to get rid of the complicated fine-grained feature matching and exhibit higher efficiency.

\section{Experiments}\label{exp}
We verify the effectiveness of our method in a series of correspondence-based tasks. We will first introduce implementation and evaluation details, and report the performance comparison with baselines. Finally, we perform detailed ablation studies for each component of our method.

\subsection{Implementation Details}\label{details}
\textbf{Backbone.} We exploit the encoder $\theta$ with ResNet-18~\cite{he2016deep}. We reduce the stride up to layer res$_4$ to get features at $1/2$ of the original image dimension for training. In our coarse-to-fine framework, we increase the stride of the encoder to 8 for coarse-grained feature matching.

\textbf{Training details.} The training is conducted on the train set of synthetic dataset FlyingThings~\cite{mayer2016large} and YouTube-VOS~\cite{xu2018youtube} collected in real-world. The FlyingThings/YouTube-VOS contains 40k/3.5k videos for training. For both datasets, we sample pair of frames which are resized into 256$\times$256 and converted to Lab color space with channel-wise dropout as the information bottleneck~\cite{lai2020mast}. The local range~$r$ in the probabilistic map is set to 24/6 for the fine-grained features~($\frac{h}{2} \times \frac{w}{2}$~(stride=2)) or coarse-grained features~($\frac{h}{8} \times \frac{w}{8}$~(stride=8)). We first train an encoder using reconstruction loss~(Eq.~(\ref{eq:self_loss})) with a batchsize of 32 for 30 epochs on FlyingThings~\cite{mayer2016large}, which is further utilized as $\theta_{self}$ in soft labeling. Then the final model $\theta$ is jointly trained for 30 epochs with proposed three losses, with a batchsize of 16 for each dataset. When training with synthetic optical flow, we filter out the points that are out of the local range or occluded by forward-backward checking. The $\tau$ is set to 1. We use Adam as the optimizer, and the initial learning rate is set to 1e-3 with a cosine~(half-period) learning rate schedule for each stage. We include more training details in the supplementary.

\begin{table*}[t]
    \centering
    \small
    \caption{\textbf{Quantitative results for point tracking on different datasets.} The frame per second (FPS) results of getting pixel-wise correspondences between a pair of video frames are measured on a single GTX-3090 at the resolution of 480$\times$640. The $^{\ast}$ indicates our coarse-to-fine framework. The - indicates the unavailable result due to the unavailable pre-trained model.} 
    \vspace{-2mm}
    \resizebox{0.8\textwidth}{!}{
       \setlength\tabcolsep{4.3pt}
       \renewcommand\arraystretch{1.00}
       \begin{tabular}{lccc|c|c|c|cc}
         \thickhline
       & &  & & BADJA & JHMDB &TAP-DAVIS & TAP-Kinetics \\
       \cline{5-9}
          \multirow{-2}{*}{Method}  & \multirow{-2}{*}{Backbone} & \multirow{-2}{*}{Stride} & \multirow{-2}{*}{FPS} & PCK@0.1$\uparrow$ & PCK@0.1$\uparrow$ & $<\delta_{avg-p}^x \uparrow$ & $<\delta_{avg-p}^x \uparrow$   \\  \hline
       
      TimeCycle~\cite{wang2019learning}  &  ResNet-50 & 8 & 28 & 41.1 & 57.3
          & 27.1 &    28.6  \\
      UVC~\cite{li2019joint}  &  ResNet-18 & 8 & 142 & 48.2 & 58.6
          & 29.0 &    25.2  \\
       CRW~\cite{jabri2020space} &  ResNet-18 & 8 & 142 & 50.9  & 59.3
       & 32.5  &    25.4  \\
       VFS~\cite{xu2021rethinking}  &  ResNet-18 & 8 & 142 & 51.9 & 60.5
       & 31.9 &  28.8  \\
      CLSC~\cite{son2022contrastive} &  ResNet-18 & 8 & 142 & - & 61.7
       & - &   -   \\
       SFC~\cite{hu2022semantic} &  ResNet-18 + ResNet-50 & 8 & 27 & 53.8 & 61.9
       & 37.2 &   31.1   \\

       MAST~\cite{lai2020mast} &  ResNet-18 & 4 & 33 & \underline{55.7} & \underline{62.4}  
       & \underline{42.5} &  \underline{33.2}  \\
       LIIR~\cite{li2022locality} &  ResNet-18 & 4 & 33 & - & 60.7
       & - &   -   \\
      \textbf{Ours$^{\ast}$} &  ResNet-18 & 8 & 34 &  \textbf{56.8} & \textbf{64.6}
      & \textbf{48.0}  & \textbf{43.8} \\
      \hdashline
      ImageNet Pre.~\cite{he2016deep} &   ResNet-18 & 2 & 8 & 57.5 & 61.5
      & 51.3 &  44.5    \\
      UVC~\cite{li2019joint} &   ResNet-18 & 2 & 8 & 56.7 & 65.1
      & 52.7 &  41.8    \\
      CRW~\cite{jabri2020space} &   ResNet-18 & 2 & 8 & 55.9  & 61.4
      & 43.2 &   37.1   \\
      VFS~\cite{xu2021rethinking} &   ResNet-18 & 2 & 8 & 58.1 & 61.0
      & 51.4 &  \underline{44.9}    \\ 
      SFC~\cite{hu2022semantic} &  ResNet-18 & 2 & 8 & 61.5 & \underline{65.9}
      & \underline{53.9} &   43.6   \\
      MAST~\cite{lai2020mast} &  ResNet-18 & 2 & 8 & \underline{63.0} & 63.1  
      & 53.8 &  42.7  \\
      \textbf{Ours} &   ResNet-18 & 2 & 8 & \textbf{67.2} & \textbf{66.8}
      & \textbf{59.8} & \textbf{48.8} \\

          \thickhline
       \end{tabular}
    }
    \captionsetup{font=footnotesize}
    \label{table:sota}
    \vspace{-4mm}
 \end{table*}

\textbf{Evaluation.}
 An important problem is how to evaluate the quality of the learned fine-grained features for pixel-wise video correspondences. Following the previous works~\cite{wang2019learning,jabri2020space,xu2021rethinking}, without any fine-tuning, we directly leverage the pre-trained encoder~$\theta$ to extract features $F_i, F_j$ with a spatial resolution of $\frac{h}{s} \times \frac{w}{s}$~(the $s$ represents the stride of the encoder) for the pair of frames $I_i, I_j$, which are later used to compute the probabilistic map $P_{i\rightarrow j}$ by Eq~(\ref{eq:local_correlation}). Based on the probabilistic map $P_{i\rightarrow j}$, we follow the recurrent inference strategy inference strategies of recent studies~\cite{wang2019learning, jabri2020space,xu2021rethinking} to propagate the target points or semantical labels of the first frame, as well as previous predictions, to the current frame $I_t$.  We evaluate the point tracking on three popular benchmarks including BADJA~\cite{biggs2019creatures}, JHMDB~\cite{jhuang2013towards}, TAP-Vid-DAVIS~\cite{doersch2022tap} and TAP-Vid-Kinetics~\cite{doersch2022tap}, and the evaluation of video object segmentation is conducted on the widely used dataset DAVIS-2017~\cite{pont20172017}.

\subsection{Results for Point Tracking}

We firstly make comparisons for point tracking since it requires the finest granularity of the learned features. The main comparators of our method are the works aim to learn good representations for matching, e.g., TimeCycle~\cite{wang2019learning}, UVC~\cite{li2019joint}, CRW~\cite{jabri2020space}, VFS~\cite{xu2021rethinking}, SFC~\cite{hu2022semantic}, and MAST~\cite{lai2020mast}. We also include the model pre-trained on ImageNet~\cite{deng2009imagenet} with human annotations. These works share a similar evaluation protocol as we mentioned in~\ref{details}. 

While previous works test the pixel-wise correspondences on JHMDB~\cite{jhuang2013towards} that only provides the human keypoint annotations. We additionally include BADJA~\cite{biggs2019creatures}, TAP-Vid-DAVIS~\cite{doersch2022tap} and TAP-Vid-Kinetics~\cite{doersch2022tap}.  We propagate the points of the first frame to other frames and evaluate the results using the annotations of each dataset. We notice some works are trained and evaluated with the coarse-grained features~(e.g., the features at 1/8  of the original image dimension). For these methods, we use our coarse-to-fine framework that also executes coarse-grained feature matching for comparisons. Meanwhile, for better comparisons,  we follow the studies in~\cite{jabri2020space,xu2021rethinking} to further reduce the stride $s$ of the encoder to 2, in order to get more fine-grained results for CRW, VFS, SFC and MAST. We also provide the FPS of computing the pixel-wise correspondences between two frames for feature-matching-based methods, which can be done by first taking the index of the maximum value in $P_{1\rightarrow2}(\cdot|i)$ for each $i$ and then applying up-sampling to get full-resolution results. More details about the evaluation are included in the supplementary material.

\textbf{Results on BADJA/JHMDB}. We adopt the standard PCK~\cite{yang2012articulated} of all visible points~(not compute PCK for each video then take average) as the evaluation metric. Each point is considered correct if it is within a distance of 0.1$\sqrt{A}$~ from the ground truth, where $A$ is the distance between keypoints~(for JHMDB) or the area of the ground-truth segmentation mask on the frame~(for BADJA). In Table~\ref{table:sota}, our method achieves 67.2\%/66.8\%, surpassing all state-of-the-art methods. Besides, our method with the coarse-to-fine design still makes the absolute improvements by 1.1\% and 2.2\% compared with MAST, and shows better efficiency compared with SFC~\cite{hu2022semantic} that uses the two-stream network to find the correspondences. More remarkably, our efficient framework even surpasses part of the methods using more fine-grained features for inference.

\begin{figure*}[t]
   \centering
   {\includegraphics[width=0.9\textwidth]{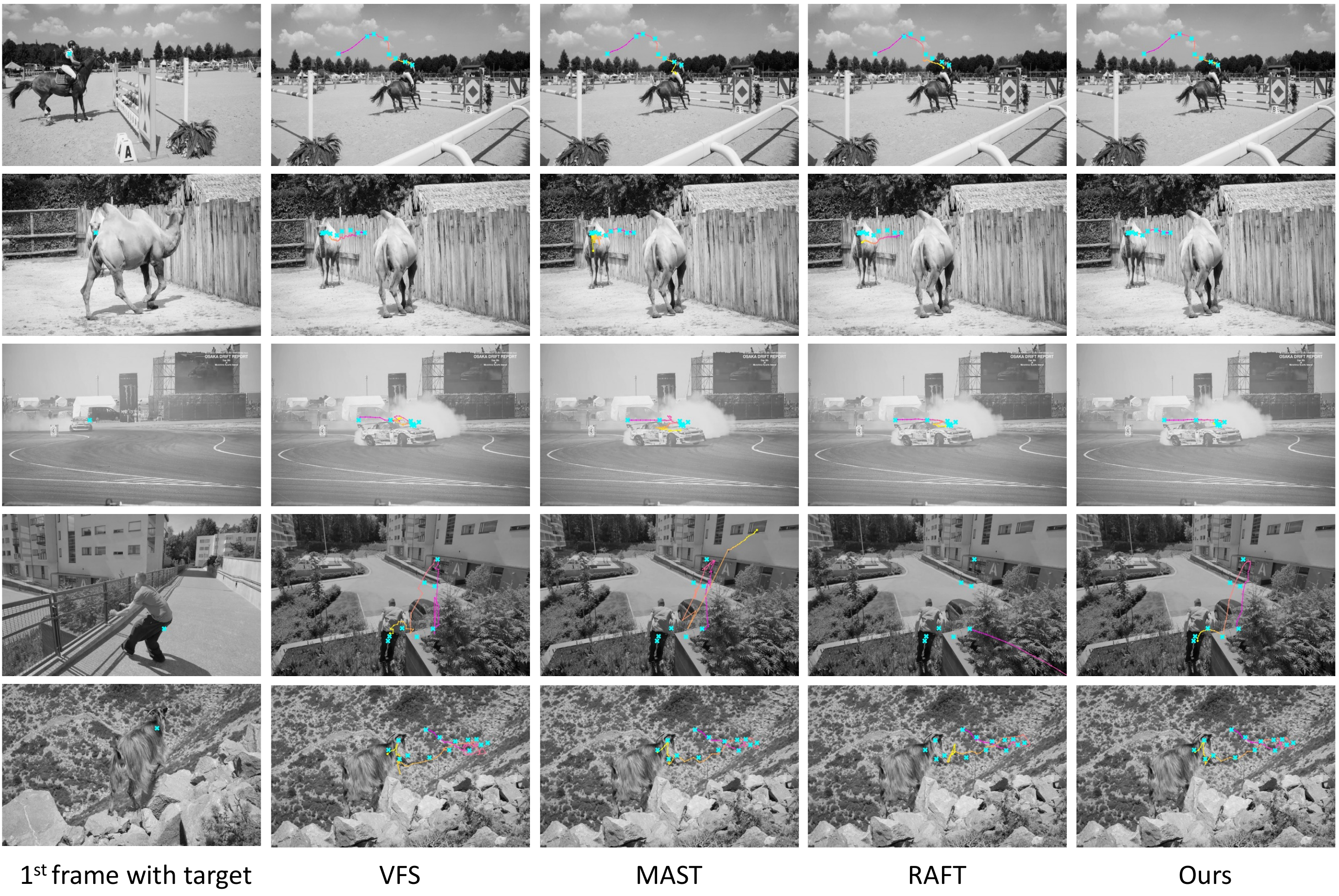}}
   \vspace{-3mm}
   \caption{\small \textbf{Qualitative results for point tracking.}  Given the target pixel in the first frame, we visualize the estimated trajectory with a pink-to-yellow colormap~(pink/yellow indicates the start/end of the video clip). We visualize sparse ground-truth labels with \textcolor{cyan}{cyan} cross marks. Please refer to the video in the supplementary for better animations. \textbf{(Zoom in for best view)}}
   \label{fig:quan}
   \vspace{-4mm}
 \end{figure*}

 \textbf{Results on TAP-Vid}. TAP-Vid is a newly developed benchmark composed of long-term videos in real-world with accurate human annotations of point tracks.  We test on the whole set of TAP-Vid-DAVIS and the test set of TAP-Vid-Kinetics. We adopt the setting of ``first fashion" in~\cite{doersch2022tap}, which tracks only into the future. The average position accuracy over all visible points~($<\delta_{avg-p}^{x}$) is adopted as the metric.  The learned fine-grained features obtain 59.8\%/48.8\% on TAP-Vid-DAVIS/TAP-Vid-Kinetics, leading apparent improvements over state-of-the-arts by 5.9\%/3.9\%. Moreover, the proposed coarse-to-fine framework gets 48.0\%/43.8\%, surpassing MAST by 5.5\%/10.6\%.

\textbf{Comparisons with task-specific methods.} Besides, we notice there are some recent methods specifically designed for point tracking, like RAFT\cite{teed2020raft}, PIPs~\cite{harley2022particle}, TAPNet~\cite{doersch2022tap}, and Thin-Slicing Net~\cite{song2017thin} even trained with human annotations. Here we also present the performance comparisons with them in Table~\ref{table:point}. It's worth noting that, to align the evaluation with these methods, except JHMDB, we first compute the $<\delta_{avg-p}^{x}$ or PCK for each video to obtain video-level results, and the final results are obtained by taking the average over all videos. Without any specific designs, our performance even surpasses these methods by 7.4\%/7.5\%/0.2\% on BADJA/TAP-DAVIS/TAP-Kinetics. 

Figure~\ref{fig:quan} shows some visualization results of the point tracking on TAP-Vid-DAVIS.  Given the target in the first frame, we visualize the estimated trajectory with a pink-to-yellow map. Compared with VFS, MAST and RAFT, our approach can output more smooth and accurate trajectories close to the sparse visualized ground truth, even facing dramatic appearance changes and deformation.

\begin{table}[t]
   \centering
   \small
   \caption{\textbf{Comparisons with methods specifically designed for point tracking.}  ``$\ddag$'' means we align the evaluation protocol with each method for fair comparisons.}
  \vspace{-2mm}
   \resizebox{0.48\textwidth}{!}{
      \setlength\tabcolsep{2pt}
      \renewcommand\arraystretch{1.00}
      \begin{tabular}{l|c|c|c|c}
        \thickhline
      &   BADJA & JHMDB & TAP-DAVIS & TAP-Kinetics  \\
      \cline{2-5}
         \multirow{-2}{*}{Method}  & PCK@0.2 $\uparrow$   & PCK@0.1 $\uparrow$ &$<\delta_{a v g}^x\uparrow$  & $<\delta_{a v g}^x\uparrow$  \\  \hline
         RAFT~\cite{teed2020raft} &  45.6 & 66.4 & 42.1 &  44.3   \\ 
         PIPs~\cite{harley2022particle}  & 62.3 & - & 55.3 &  48.2   \\
         TAPNet~\cite{doersch2022tap}  & -  &  62.3 & 48.6  & 54.4      \\
         Thin-Slicing Net~\cite{song2017thin}  & -  &  \textbf{68.7} & -  & -      \\
     \hline
     \textbf{Ours}$^{\ddag}$  & \textbf{69.7} & 66.8 &\textbf{62.8} &    \textbf{54.6}   \\
         \thickhline
      \end{tabular}
   }
   \captionsetup{font=footnotesize}
   \vspace{-5mm}
   \label{table:point}
\end{table}

\begin{table}[t]
    \centering
    \small
    \caption{\textbf{Quantitative results for video object segmentation on DAVIS$_{17}$}~\cite{pont20172017}. ``Sup.'' means using human annotations for training.}
    \vspace{-3pt}
    \resizebox{0.48\textwidth}{!}{
       \setlength\tabcolsep{3.3pt}
       \renewcommand\arraystretch{1.00}
       \begin{tabular}{lcc|ccc}
         \thickhline
       & &  &  \multicolumn{3}{c}{DAVIS$_{17}$}  \\
       \cline{4-6}
          \multirow{-2}{*}{Method} & \multirow{-2}{*}{Sup.} & \multirow{-2}{*}{Backbone}  & $\mathcal{J}$\&$\mathcal{F}_m\uparrow$     &  $\mathcal{J}_m\uparrow$ & $\mathcal{F}_m\uparrow$  \\  \hline
          TimeCycle~\cite{wang2019learning} & & ResNet-50  & 40.7 & 41.9 & 39.4       \\
          UVC~\cite{li2019joint} & & ResNet-18  & 59.5 & 57.7 & 61.3        \\
         MAST~\cite{lai2020mast} && ResNet-18     & 65.5
         &  63.3 & 67.6    \\
         CRW~\cite{jabri2020space} && ResNet-18     & 67.6
         & 64.8 & 70.2     \\
         JSTG~\cite{zhao2021modelling} &&  ResNet-18 
         & 68.7 & 65.8   & 71.6  \\
         VFS~\cite{xu2021rethinking}  && ResNet-50   & 68.9 & 66.5 & 71.3    \\
         DUL~\cite{araslanov2021dense} && ResNet-18  &  69.3 
         & 67.1 & 71.6     \\
         MAMP~\cite{miao2022self} && ResNet-18    & 69.7
         &  68.3 & 71.2    \\

         CLTC~\cite{jeon2021mining} && ResNet-18 & 70.3 & 67.9& 72.6\\
          CLSC~\cite{son2022contrastive} &&  ResNet-18  
          & 70.5 & 67.4   & 73.6  \\
          SFC~\cite{hu2022semantic} && ResNet-18 + ResNet-50     & 71.2
          & 68.3 & 74.0      \\

          LIIR~\cite{li2022locality} && ResNet-18     & 72.1
          & 69.7 & 74.5    \\
     
      \hline
       \textbf{Ours} &&  ResNet-18   &\textbf{72.4} & \textbf{70.5}
          & \textbf{74.4}   \\

      \hline

      OSVOS-S~\cite{maninis2018video} & $\checkmark$& VGG-16     & 68.0
      & 64.7 & 71.3    \\
      FEELVOS~\cite{voigtlaender2019feelvos} &$\checkmark$ & Xception-65     & 71.5
      & 69.1 & 74.0    \\
          \thickhline
       \end{tabular}
    }
    \captionsetup{font=footnotesize}
    \label{table:vos_sota}
    \vspace{-7mm}
 \end{table}

\subsection{Results for Video Object Segmentation}
\begin{figure}[t]
   \centering
   {\includegraphics[width=0.47\textwidth]{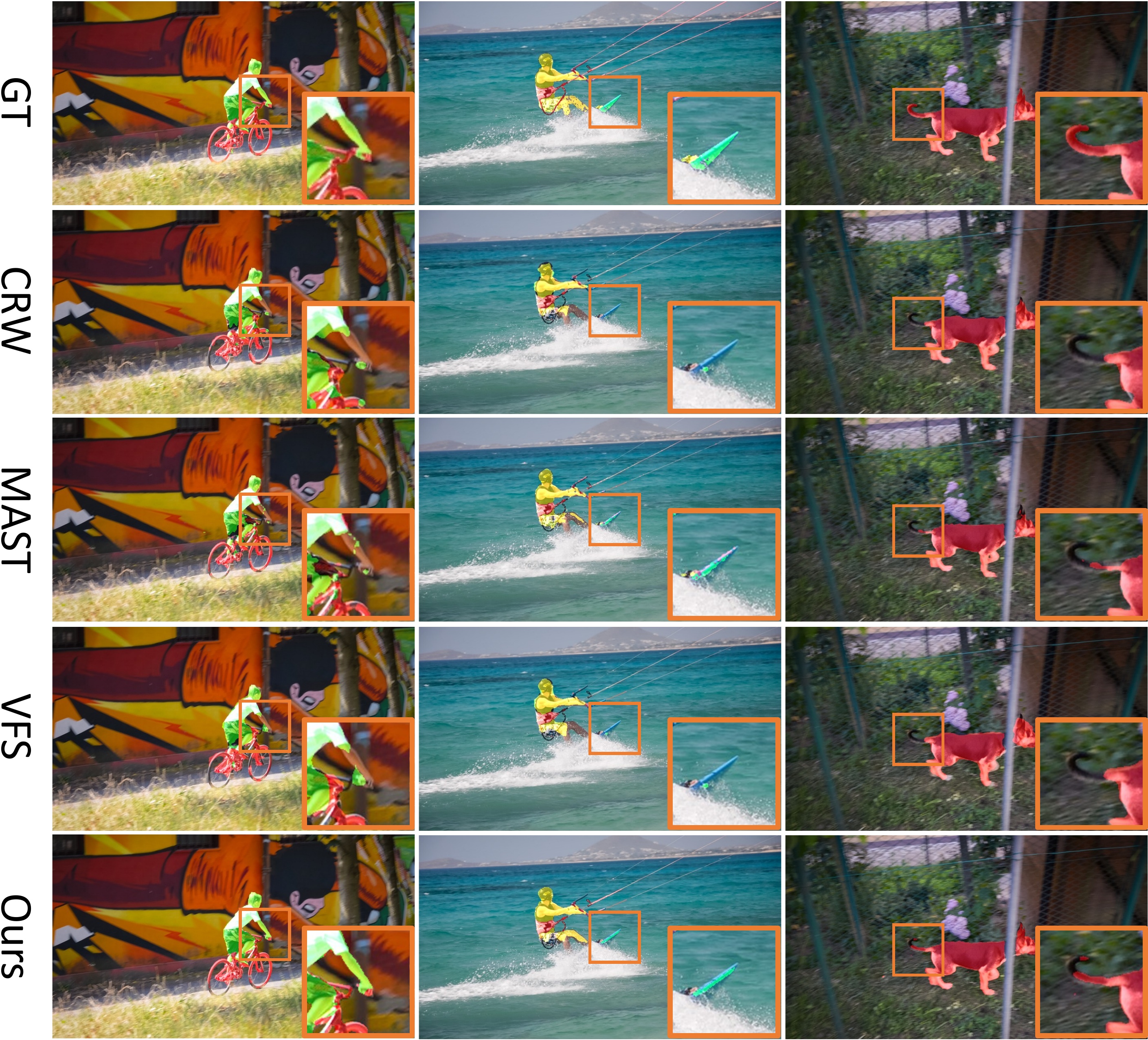}}
   \caption{\small \textbf{Qualitative results for video object segmentation.}  Given the semantic mask in the first frame, we show the propagation results on the target frame. (\textbf{Zoom in for best view})}
   \label{fig:quan_vos}
   \vspace{-5mm}
\end{figure}

Next, we evaluate methods with semi-supervised video object segmentation. We use the mean of region similarity $\mathcal{J}_m$, mean of contour accuracy $\mathcal{F}_m$ and their average $\mathcal{J} \& \mathcal{F}_m$ as the evaluation metrics. In Table~\ref{table:vos_sota}, our method still leads the performance. More remarkably, our method even outperforms some task-specific fully-supervised algorithms~\cite{maninis2018video, voigtlaender2019feelvos}. Here we select several representative videos for inference, and give the visualization results in Figure~\ref{fig:quan_vos}, our method produces tight boundaries around the object areas, and obtains more fine-grained results, especially for small objects. For example, in the first column of Figure~\ref{fig:quan_vos}, the tiny arm of the human can still be segmented, which further demonstrates the advantages of learning fine-grained features for video correspondences. However, we find that fine-grained features may hinder object-centric feature learning since it may rely more on low-level patterns (e.g., texture, color, etc), which degrades the performance in video object segmentation to some extent. We regard it as our future work.

 \subsection{Ablation Study}
 We perform ablation study with point tracking on the TAP-Vid-DAVIS~\cite{doersch2022tap} dataset.

\textbf{Learning with synthetic videos.} We study the effect of three different ways as defined in Eq.~(\ref{eq:klv1}), Eq.~(\ref{eq:klv2}) and Eq.~(\ref{eq:klv3}) for learning the fine-grained features with synthetic videos. Table~\ref{tab:abalation_kl} shows the performance comparisons across the three kinds of objective functions. As indicated by the results, the loss~$\mathcal{L}_{\text{KL-v1}}$ using dirac distribution to generate the labels performs badly. We think the deterministic labels were given on synthetic videos, resulting in the learned features not being robust enough on real-world videos. Besides, we find introducing the gaussian distribution in~$\mathcal{L}_{\text{KL-v2}}$ results in performance degradation. The performance drops a lot when progressively increasing the variance, which may be attributed to the inability of gaussian distribution to approach the real probabilistic distribution since it is extremely complicated.  Expectedly, the proposed soft labeling boosts up the performance from 42.6\% to 55.8\% when only pre-training the $\theta_{self}$ with Eq.~(\ref{eq:self_loss}). Moreover, we also try another 2D encoder pre-trained with the contrastive loss $\mathcal{L}_{\text{nce}}$~\cite{he2020momentum} on ImageNet~\cite{deng2009imagenet}, which has a stronger ability to capture the spatially discriminative features. The results are further improved to 57.5\%, which motivates us to leverage a more powerful 2D feature extractor for obtaining the soft labels. We regard it as future work.


\begin{table}[]
   \centering
   \captionsetup{font=small}
   \caption{\textbf{Ablation study for $\mathcal{L}_{\text{KL}}$}. The $\sigma$~/~$\theta_{self}$ represents the variance~/~encoder used in $\mathcal{L}_{\text{KL-v2}}$~/~$\mathcal{L}_{\text{KL-v3}}$.} 
   \vspace{-2mm}
   \resizebox{0.4\textwidth}{!}{%
   \setlength\tabcolsep{4pt}
   \begin{tabular}{c|c|c|c}
   \thickhline
   \multirow{2}{*}{Obj. Function} & \multicolumn{2}{c|}{Hyper-param.} & TAP-Vid-DAVIS  \\ 
   \cline{2-3}
                                  &            $(\sigma_{u},\sigma_{v})$       &   $\theta_{self}$  & $<\delta_{a v g}^x \uparrow$ \\ \hline
   $\mathcal{L}_{\text{KL-v1}}$               & -  & -         &         42.6        \\ \hline
   \multirow{3}{*}{$\mathcal{L}_{\text{KL-v2}}$} & (1,1)  & -         &       41.9              \\
                             & (3,3)  & -         &       33.8               \\
                             & (6,6) & -         &         29.7             \\ \hline
   \multirow{2}{*}{$\mathcal{L}_{\text{KL-v3}}$}  & -  &      w.~$\mathcal{L}_{\text{rec}}$ & 55.8 \\ 
                             & -  &      w.~$\mathcal{L}_{\text{nce}}$~\cite{he2020momentum} & \textbf{57.5} \\
   \thickhline
   \end{tabular}%
   }
   \vspace{-2mm}

   \label{tab:abalation_kl}
\end{table}

\begin{table}
   \centering
   \small
   \captionsetup{font=small}
   \caption{\textbf{Ablation study for training objective functions.} FT: FlyingThings~\cite{mayer2016large}. YTV: YouTube-VOS~\cite{xu2018youtube}.}
   \vspace{-2mm}
   \resizebox{0.4\textwidth}{!}{
     \setlength\tabcolsep{4pt}
       \begin{tabular}{ccc|c|c}
       \thickhline
         \multirow{2}{*}{$\mathcal{L}_{\text{KL}}$}& \multirow{2}{*}{$\mathcal{L}_{\text{rec}}$} &  \multirow{2}{*}{$\mathcal{L}_{\text{adv}}$}& \multirow{2}{*}{Training Data} & TAP-Vid-DAVIS  \\ 
          &  &   &  & $<\delta_{a v g}^x \uparrow$ \\ \hline
         $\checkmark$ &  & & FT & 55.8 \\
          & $\checkmark$ &   & YTV & 56.4 \\
          & $\checkmark$ &   & FT + YTV & 56.7 \\

         $\checkmark$ & $\checkmark$ &  & FT + YTV&   59.2  \\
         $\checkmark$ & $\checkmark$ & $\checkmark$ & FT + YTV& \textbf{59.8}  \\
       \thickhline
     \end{tabular}
   }
   \vspace{-6mm}
   \label{tab:abalations_temporal}
\end{table}

 \textbf{Different training objective functions.} We examine how different objective functions impact the overall performance, which is shown in Table~\ref{tab:abalations_temporal}. The $\mathcal{L}_{\text{KL}}$, $\mathcal{L}_{\text{rec}}$ and $\mathcal{L}_{\text{adv}}$ denotes the losses defined in Eq.~(\ref{eq:klv3}), Eq.~(\ref{eq:self_loss}) and Eq.~(\ref{eq:da_loss}). Surprisingly, we find the model trained on unlabeled real-world videos with $\mathcal{L}_{\text{rec}}$ obtains a better result, which indicates better generalization of self-supervised feature learning. By leveraging both $\mathcal{L}_{\text{KL}}$ and $\mathcal{L}_{\text{rec}}$ with synthetic and real-world videos, the performance is further improved to 59.2\%.  As expected, executing $\mathcal{L}_{\text{adv}}$ to address the domain mismatch improves performance by 0.6\%. By fusing three losses, the performance reaches 59.8\%. The results consistently indicate that incorporating self-supervised and adversarial training with unlabeled data exhibits a performance boost against the model only trained with synthetic data.

\vspace{-2mm}

\section{Conclusions}
In this paper, we address pixel-wise video correspondences by learning fine-grained features. We propose to use not only labeled synthetic videos but also unlabeled real-world videos for feature learning. We first study how to take advantage of synthetic supervision for feature learning, and we find directly utilizing the motion vector results in degradation for the learned features. Thus, we propose soft labeling to address the issue. To improve the generalization, we introduce self-supervised reconstructive learning into the overall training and further enhance the features by leveraging adversarial training. Moreover, we propose a coarse-to-fine framework to alleviate the problem of computational efficiency. Extensive experiments on the downstream tasks validate the effectiveness of the proposed feature learning method and our efficient design.

{\small
\bibliographystyle{ieee_fullname}
\bibliography{egbib}
}

\end{document}